\documentclass[sigconf]{acmart}
\usepackage{balance}
\usepackage{booktabs} % For formal tables
\usepackage{subcaption}
\usepackage{soul}

\copyrightyear{2019}
\acmYear{2019}
\setcopyright{rightsretained}
\acmConference[KDD '19]{The 25th ACM SIGKDD Conference on Knowledge Discovery and Data Mining}{August 4--8, 2019}{Anchorage, AK, USA}
\acmBooktitle{The 25th ACM SIGKDD Conference on Knowledge Discovery and Data Mining (KDD '19), August 4--8, 2019, Anchorage, AK, USA}
\acmDOI{10.1145/3292500.3330766}
\acmISBN{978-1-4503-6201-6/19/08}

\acmSubmissionID{ads1642p}

\begin{document}
\fancyhead{}

% The "title" command has an optional parameter, allowing the author to define a "short title" to be used in page headers.
\title[Seasonal-adjustment Based Feature Selection Method \\ for Large-scale Search Engine Logs]{Seasonal-adjustment Based Feature Selection Method \\ for Predicting Epidemic with Large-scale Search Engine Logs}

% The "author" command and its associated commands are used to define the authors and their affiliations.
% Of note is the shared affiliation of the first two authors, and the "authornote" and "authornotemark" commands
% used to denote shared contribution to the research.
\author{Thien Q. Tran}
\email{thientquang@mdl.cs.tsukuba.ac.jp}
\affiliation{%
  \institution{University of Tsukuba}
  \streetaddress{Tennodai 1--1--1}
  \city{Tsukuba}
  \state{Ibaraki}
  \country{Japan}
  \postcode{305-8571}
}

\author{Jun Sakuma}
\email{jun@cs.tsukuba.ac.jp}
\affiliation{%
  \institution{University of Tsukuba, Riken AIP}
  \streetaddress{Tennodai 1--1--1}
  \city{Tsukuba}
  \state{Ibaraki}
  \country{Japan}
  \postcode{305-8571}
}

%
% The abstract is a short summary of the work to be presented in the article.
\begin{abstract}
  Search engine logs have a great potential in tracking and predicting outbreaks of infectious disease. More precisely, one can use the search volume of some search terms to predict the infection rate of an infectious disease in nearly real-time. However, conducting accurate and stable prediction of outbreaks using search engine logs is a challenging task due to the following two-way instability characteristics of the search logs. First, the search volume of a search term may change irregularly in the short-term, for example, due to environmental factors such as the amount of media or news. Second, the search volume may also change in the long-term due to the demographic change of the search engine. That is to say, if a model is trained with such search logs with ignoring such characteristic, the resulting prediction would contain serious mispredictions when these changes occur.
 
  In this work, we proposed a novel feature selection method to overcome this instability problem. In particular, we employ a seasonal-adjustment method that decomposes each time series into three components: seasonal, trend and irregular component and build prediction models for each component individually. We also carefully design a feature selection method to select proper search terms to predict each component. We conducted comprehensive experiments on ten different kinds of infectious diseases. The experimental results show that the proposed method outperforms all comparative methods in prediction accuracy for seven of ten diseases, in both now-casting and forecasting setting. Also, the proposed method is more successful in selecting search terms that are semantically related to target diseases.
\end{abstract}

%
% Keywords. The author(s) should pick words that accurately describe the work being
% presented. Separate the keywords with commas.

%
% This command processes the author and affiliation and title information and builds
% the first part of the formatted document.
\maketitle
\section{Introduction and background}
Search engine logs have a great potential in tracking and predicting infectious disease outbreaks \cite{brownstein2009digital}. Search logs are accessible in nearly real-time and can be used for detecting epidemic outbreaks more quickly than gathering data from hospitals nationwide, which often takes one or two weeks. A number of studies have been conducted using search logs of different search engines aiming at prediction of epidemic outbreak, such as Google \cite{ginsberg2009detecting}\cite{copeland2013google}\cite{lampos2015advances}, Yahoo \cite{polgreen2008using} and Baidu \cite{yuan2013monitoring}. These studies selected a number of search terms and then learned a prediction model in which the search rates of selected search terms were used as the features. Some of these works showed very promising results. For example, \cite{ginsberg2009detecting} reported a prediction accuracy with the correlation of 0.97 in the outbreak prediction problem for influenza.

Conducting accurate prediction using search engine logs is still a challenging task. The first difficulty is the instability of search log data. The search volume of a search term may change in both the short-term and the long-term due to many factors. First, information retrieval activities of users are very unstable and easily affected by environmental factors such as the amount of media or news. Thus, the search volume of a search term may irregularly fluctuate in the short-term \cite{butler2013google}. Second, the search volume of a search term may also change gradually in the long run, for example, due to the demographic change of search engine \cite{monica2017tech}. Without considering such short-term and long-term changes, the resulting prediction would contain serious overestimation or underestimation.

Another difficulty is the computational hardness due to the enormously large number of search terms. For example, on Yahoo! Japan search engine, about 30 millions of different search terms are searched a day, on average. This makes the dimension of the explanatory variable very large. To reduce the learning cost and  overfitting, one needs to conduct feature selection to select effective search terms for the prediction model \cite{guyon2003introduction}.

Moreover, in the epidemic prediction problem, a reliable predictor not only needs to predict outbreaks accurately but also needs to produce the prediction based on meaningful evidence. More precisely, it is necessary to select search terms semantically related to the target infectious disease so that the resulting model is interpretable to experts. Finding such terms is difficult especially for infectious diseases with strong seasonality. Because the outbreak of these seasonal epidemics occurs at around the same time every year, some seasonal search terms not related to the target disease frequently co-occurs with the outbreak. While the total number of the search terms is large, only a small number of search terms are semantically related to the target disease and are buried in non-related search terms. This makes these related search terms extremely hard to find.

Finally, because search logs have a great potential in monitoring various epidemic diseases based on its diversity, methodologies that are applicable to a wide variety of infectious diseases are preferable. However, all existing works only focused on outbreak prediction of one or two major infectious diseases (e.g., influenza and dengue). This left the question of the applicability to other infectious diseases.

In breif, we aim to develop a feature selection method that is:
\begin{itemize}
\item \textbf{Stable}. The selected search terms must be robust against the instability of search log data, i.e, the changes in both short-term and long-term.
\item \textbf{Scalable}. It must be implementable and be executable in a reasonable time for large-scale search log data, e.g, up to millions of search terms.
\item \textbf{Interpretable}. It is desirable that the selected search terms are semantically related to the target infectious disease.
\item \textbf{Universal}. It must be applicable to a wide range of infectious diseases.
\end{itemize}

In this work, we introduce a novel feature selection method that takes the instability characteristic of the search logs into account. Our method is built on a decomposition prediction framework. Our first insight is that because the short-term and the long-term changes have different occurrence dynamics, it is better to deal with these changes separately to conduct better prediction. Thus, we decompose these time serieses into three components: seasonal component, trend component and irregular component. Then, we individually train models to predict each component. The final prediction of the infection rate is obtained by combining the prediction of each component. Here, the trend and irregular component can be interpreted as the long-term and short-term change, respectively.

Moreover, we do not only train individual models but also conduct feature selection individually for each component. This is because useful search terms for prediction can vary for each component. Thus, we carefully design a method to select proper search terms that can appropriately predict each component. Furthermore, because we consider the seasonal component separately, our method can avoid selecting non-related seasonal search terms that are highly correlated with the outbreak. Hence, as we show with experimental results later, the resulting models built with our method are also more interpretable.

We note that, in the prediction problem, model selection is an orthogonal task to the feature selection problem. We emphasize that this work focuses on developing a feature selection method, and at some point, a prediction modeling framework for selected search terms. That is, various prediction models can be readily applied to our method. However, because most of the existing studies of search terms selection evaluated their methods with a linear prediction model, we only adopt a simple multivariate linear regression model as the prediction model for the evaluation purpose.

We conducted comprehensive experiments on predicting outbreaks of ten different kinds of infectious diseases to demonstate the universality of the proposed method. From the experimental results, we show that, the prediction model learned with the search terms selected by the proposed methods achieved more accurate and stable prediction compared to the existing methods. In particular, the proposed method achieved better prediction accuracy compared to all comparative methods for seven of ten diseases. For example, in the case of herpangina, we achieved the correlation of $0.98$, which is higher than the correlation coefficient of other methods by at least 12\%. We also achieved a prediction error that is at least 2 times smaller than other methods. Moreover, our method outperformed the existing methods in both the now-casting and the forecasting setting. Furthermore, not only in prediction accuracy, our method was also more successful in selecting search terms that are semantically related to the target diseases. Especially, the proposed method was able to select related search terms even in the forecasting setting while the comparative methods almost failed.

\section{Related works} \label{related-work}
There are some prediction models have been proposed for the outbreak prediction problem. For example, the GFT method used a univariate logit regression model in which the single aggregate variable obtained by summing up selected search terms were used as the feature \cite{ginsberg2009detecting}. On the other hand, some works employed more representative models, for example, the multivariable regression model, and achieved a better prediction accuracy compared to the GFT model \cite{copeland2013google}\cite{lampos2015advances}. Besides, some studies also proposed using the nonlinear frameworks, such as the Gaussian Processes (GPs)\cite{lampos2015advances}\cite{lampos2017enhancing}.

Several approaches to select the search terms has been proposed. Some works attempted to leverage prior knowledge about the target disease to select relevant search terms. For example, some studies preliminarily chose keywords related to the target disease and then used these keywords for filtering the search terms \cite{yuan2013monitoring}\cite{polgreen2008using}. There is also an attempt that applies text processing technique to determine meaningful search terms \cite{lampos2017enhancing}. One common point of these methods is that one has to preliminarily prepare keywords related to the target infectious disease manually. This task is often difficult because of the following two reasons. First, the search terms in search logs are extremely diverse and it is hard to cover all possible keywords (e.g, due to the user's diversity, synonyms, etc.). Secondly, finding related keywords could be hard for some infectious diseases or be costly when conducting for multiple diseases.

There also have been some efforts to automatically select useful search terms without prior knowledge. The Google Flu Trends method (the GFT method) used a correlation based method with the hybrid feature selection framework \cite{ginsberg2009detecting}. Specifically, they first learned a logit regression model for each search term and then ranked all the search terms by the correlation coefficients between the predicted value and the ground-truth. Finally, they select the best feature subset from the search terms with the top-rank scores and obtain a single aggregate variable by summing up the search rates of selected search terms. Despite the fact that the GFT method is scalable and selecting features using correlation coefficient is effective in collecting related search terms, \cite{copeland2013google} reported that the aggregated feature performs poorly in the prediction task.

Another approach is to use sparse modeling with the regularized regression schemes, e.g, the LASSO \cite{tibshirani1996regression} method which uses the L1 regularization and the Elastic Net method which combines both the L1 and L2 regularization \cite{zou2005regularization}. Based on the observation that many search terms have search rate time serieses that are strongly correlated to each other, \cite{lampos2015advances} used the Elastic Net method because it is known to perform well in multicollinearity setting. The Elastic Net method can select useful search terms for prediction and can learn a sparse model which is helpful in understanding the model. However, there is no guarantee that the selected search terms are related to the target disease. Also, learning with the Elastic Net method could be costly for large-scale search logs, e.g, due to the high optimization and hyperparameters tuning cost.

There are also studies focused on the seasonal characteristic of the infectious epidemic. In particular, \cite{dalum2017seasonal} extended the GFT model by introducting a different ranking rule. To be specific, they first modeled the seasonal variation of the observed time series of the infection rate using the Serfling model and the yearly average model. After that, they ranked the search terms by the correlation of the search rate time serieses with the residual of the seasonal model and the observed infection rate. The final subset of search terms is selected from the ranked list. However, in their experiments because they used only 100 terms obtained from Google Correlate for evaluation, applicability to large-scale search logs is questionable. Our experimental result using their method with our large-scale dataset indicated that their method failed to select proper search terms due to the extreme diversity of the large-scale search logs.

Above all, there was no study that took changes in short-term and long-term in the search logs into account. Moreover, all existing works only focused on one or two specific infectious diseases. This left the question of the applicability to other infectious diseases. In conclusion, none of these methods fully achieved our goal.

\section{Data description}
In this section, we describe the data used in this work.

\vspace{.5em}
\textbf{Epidemic report data}: We use the publicly available Infectious Diseases Weekly Report (IDWR) data published by the National Institute of Infectious Diseases (Japan) as the ground-truth data for epidemic breakout \cite{idwr-data}. IDWR data contains the number of the infection cases for all infectious diseases with the reporting requirement in Japan. We denote the infection rate of an infectious disease by $y^d_t$, where $d$ is the disease name and $t$ is a weekly time. Here, the infection rate $y_t$ is the number of the infection cases reported in IDWR data normalized by the population of Japan at time $t$. While IDWR data is published every week, the data of the epidemic at time $t$ is published one week later, i.e., at time $t+1$ because it takes time for gathering the data from the hospitals. Among the infectious diseases reported in IDWR data, we focus on ten infectious diseases listed in Table \ref{table:result}. Here we chose ten common diseases which has a relatively large number of infected cases to make sure that desired evidences shows up in our data search logs. Targeting at ten diseases is significantly different from the existing works since they only focused on one or two major infectious diseases.

\vspace{.5em}
\textbf{Search log data:}
For the search log data, we use the search log data of Yahoo! JAPAN search engine. Currently, Yahoo! JAPAN has approximately 120 million active users quarterly, occupying most of the population of Japan. Therefore, one can use this data to observe the interest of the people in Japan in nearly real-time. Yahoo! JAPAN search log data is collected daily, containing all the search terms in that day and the search volume of each search term. In our work, instead of the search volume, we use the search rate, which is obtained as the search volume of a search term normalized by the search volume of all search terms at that moment. We denote the search rate of search term $q^k$ at time $t$ as $x^k_t$ and the data of all search terms as $X^k_t$.

\vspace{.5em}
\textbf{Training and test data}:
To simplify the notation, we focus on a specific target disease and omit the subscription $d$ - disease name. We define the data set of all search terms and the epidemic data as $D = \{(X_t, y_t)\}$. We then set $80\%$ of $D$ for the training data and the remaining data is used for the test data. We used the training data for the feature selection and model learning. The test data is used only for evaluating the performance of the final model.

\section{Problem formulation}
\begin{figure*}[t]
  \centering
  \includegraphics[width=.95\textwidth]{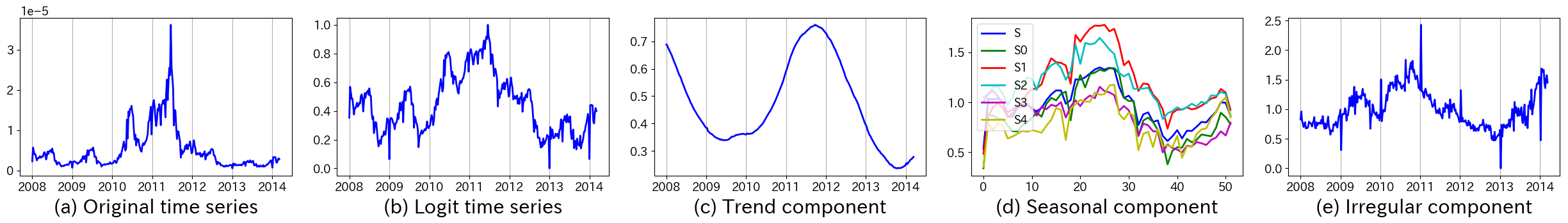}
  \vspace{-0.5em} 
  \caption{Decomposition result of the infection rate time series for erythema infectiosum\label{fig:decomposite}}
  \vspace{-0.5em}
\end{figure*}
In this section, we will formulate the epidemic prediction problem using the search log data.

\vspace{.5em}
\textbf{Objective variable}: The value to predict is the infection rate $y_t \in [0,1]$ at time $t$. We describe the time series of infection rate (ground-truth) as $Y = \{y_1, y_2, \dots, y_T\}$ where $y_t$ denotes the infection rate $y_t$ at time $t$.

\vspace{.5em}
\textbf{Explanatory variable}: We use $N$ time serieses $X = \{X^1, X^2, \dots$ $, X^N\}$ of $N$ search terms $q^1, q^2, \dots, q^N$ as the explanatory variables. Here, $X^k=\{x^k_1, x^k_2, \dots, x^k_T\}$ is the time series of search term $q^k$, and $x^k_t\in[0,1]$ is the search rate of search term $q^k$ at time $t$. We also denote the search rate of search term $q^k$ from time $1$ to time $t$ as $X^k_{1:t}$ and $X_{1:t} = \{X^1_{1:t}, X^2_{1:t}, \dots, X^N_{1:t}\}$ for all the search terms.

\vspace{.5em}
\textbf{Prediction problem}: The prediction problem is the problem of learning a function $f$ that takes the search log data available until time $t$, $X_{1:t}$ as the input and returns the infection rate $y_{t+\phi}$ at time $t + \phi$. When $\phi=0$, we predict the current epidemic at that moment, rather than the future epidemic.

\section{Proposed method}
Here, we discuss the overall framework and the details of our proposed method. First, our overall framework is described as follows:

\begin{enumerate}
\item \textbf{Seasonal adjustment:} We decompose the logit time series $logit(Y)$ of the epidemic into three components: trend $\hat{T}$, seasonal $\hat{S}$, and irregular component $\hat{I}$. Similarly, we also decompose the logit time series of 2.5 million candidate search terms. To be specific, for the search term $q^k$, we decompose $logit(X^k)$ into $\hat{T}^k$, $\hat{S}^k$ and $\hat{I}^k$.
\item \textbf{Feature ranking:} Next, we score the prediction capability of the trend component $\hat{T}^k$ and the irregular component $\hat{I}^k$ of each search term individually. We then sort all the search terms using these scores in descending order.
\item \textbf{Feature subset selection:} From the sorted search term list, we find the optimal combination of search terms for predicting the trend component $\hat{T}$, and irregular component $\hat{I}$ using the wrapper approach for feature subset selection \cite{kohavi1997wrappers}. We apply the forward selection method to choose the candidate search terms and use the multivariate regression model with L2 regularization for the prediction model.
\item \textbf{Prediction model:} Finally, we retrain the prediction model $f_{\hat{T}}$ for the trend component $\hat{T}$, and $f_{\hat{I}}$ for the irregular component $\hat{I}$ with the selected search terms.
\end{enumerate}

\subsection{Seasonal adjustment}
Seasonal adjustment is the method for excluding the seasonal component for seasonal time series. Considering that we need to process numerous search terms, we use a basic seasonal adjustment method with the multiplicative model $logit(y_t) = \hat{T}_t * \hat{S}_t * \hat{I}_t$ where the logit time series $logit(Y)$ is decomposed into three components: trend $\hat{T}$, seasonal $\hat{S}$, and irregular component $\hat{I}$. Similarly for the search term $q^k$, we decompose $logit(X^k)$ into $\hat{T}^k$, $\hat{S}^k$ and $\hat{I}^k$. Next, we describe the interpretations of these components in the epidemic prediction problem and their calculation methods.

As a running example, we use the infection rate of erythema infectiosum. We show the original time series and the logit time series of infection rate of erythema infectiosum in Figure \ref{fig:decomposite}(a) and Figure \ref{fig:decomposite}(b), respectively.

\subsubsection{\textbf{Trend component $\hat{T}$}}\

The trend component represents the variation of the outbreak in the entire year, i.e., the long-term change in the time series. The trend component is calculated by the moving average. Because most of the infectious diseases focused in our work have a one-year cycle, we use the moving average ${MA}_{52}$ with the width of 52 weeks (see Figure \ref{fig:decomposite}c). Because we used moving average of the past 52 weeks, data in the first year becomes unusable
\begin{equation}
  \hat{T}_t={MA}_{52}(logit(y_t))=\frac{1}{52}\sum_{i=t-52}^{52}logit(y_i).
\end{equation}

\subsubsection{\textbf{Seasonal component $\hat{S}$}}\

The seasonal component $\hat{S}$ represents the common pattern of each cycle and is calculated as follows. First, we calculate the detrended time series $\hat{D_t}=\frac{logit(y_t)}{\hat{T}_t}$ by removing the trend component from the original time series. The seasonal component $\hat{S}$ is calculated as the average of all cycles of $\hat{D}$. Specifically, if we denote the set of indexes of data points which have the same position with respect to the cycle as $I_t=\{\dots,t-2m,t-m,t,t+m,t+2m,\dots\}$, the seasonal component $\hat{S}$ is derived as follows (see Figure \ref{fig:decomposite}d)
\begin{equation}
  \hat{S}_t=\frac{1}{|I_t|}\sum_{i\in I_t}\hat{D}_i.
\end{equation}

We assume that the seasonal component does not change in a short time. For this reason, we do not build a prediction model for the seasonal component, but instead, use the seasonal pattern $\hat{S}$ calculated by the epidemic data in the training period for prediction.

\begin{figure*}[t]
  \centering
  \begin{subfigure}{.3\textwidth}
    \centering
    \includegraphics[width=1\textwidth]{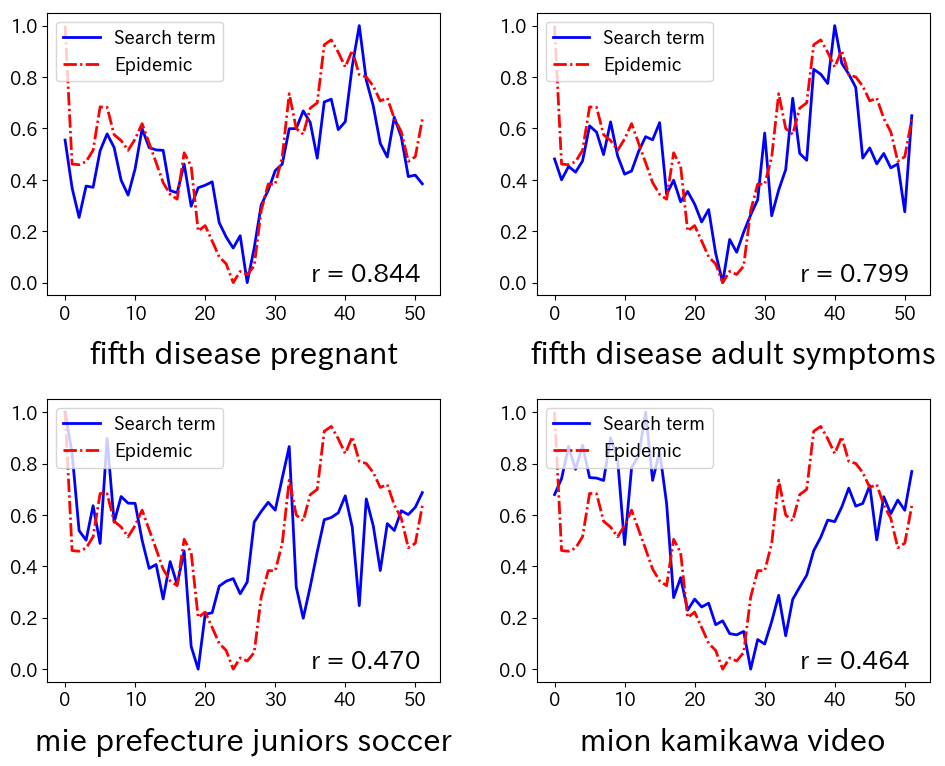}
    \caption{Season component: $\hat{S}$ and $\hat{S}^k$}
  \end{subfigure}%
  \begin{subfigure}{.3\textwidth}
    \centering
    \includegraphics[width=1\textwidth]{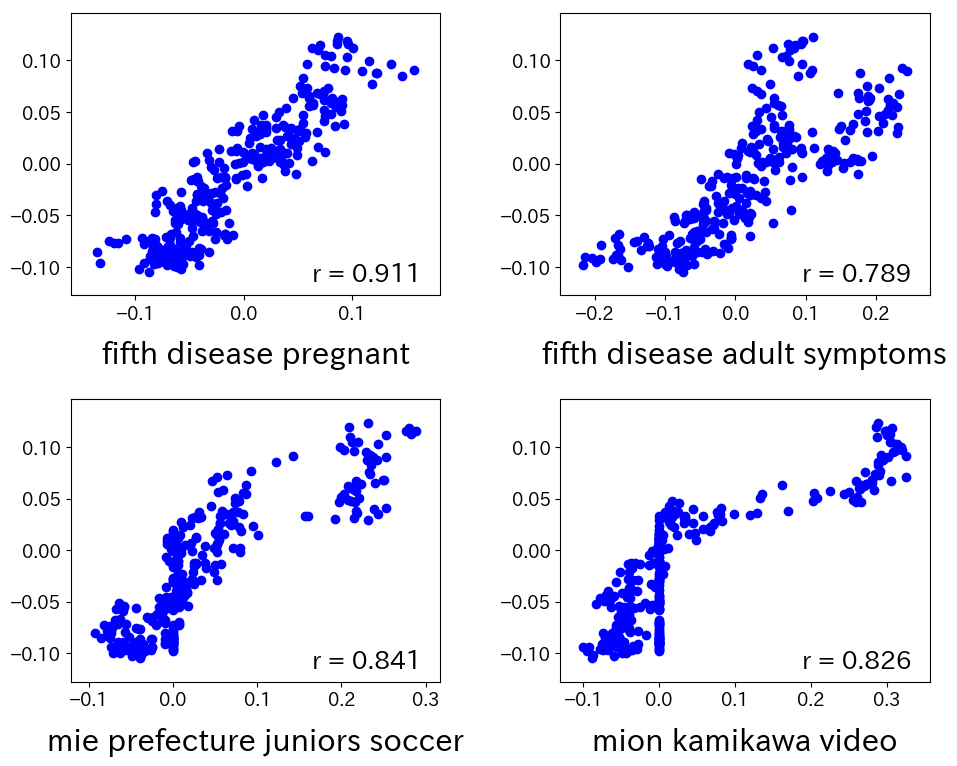}
    \caption{Trend component: $\Delta_1(\hat{T})$ and $\Delta_1(\hat{T}^k)$}
  \end{subfigure}%
  \begin{subfigure}{.3\textwidth}
    \centering
    \includegraphics[width=1\textwidth]{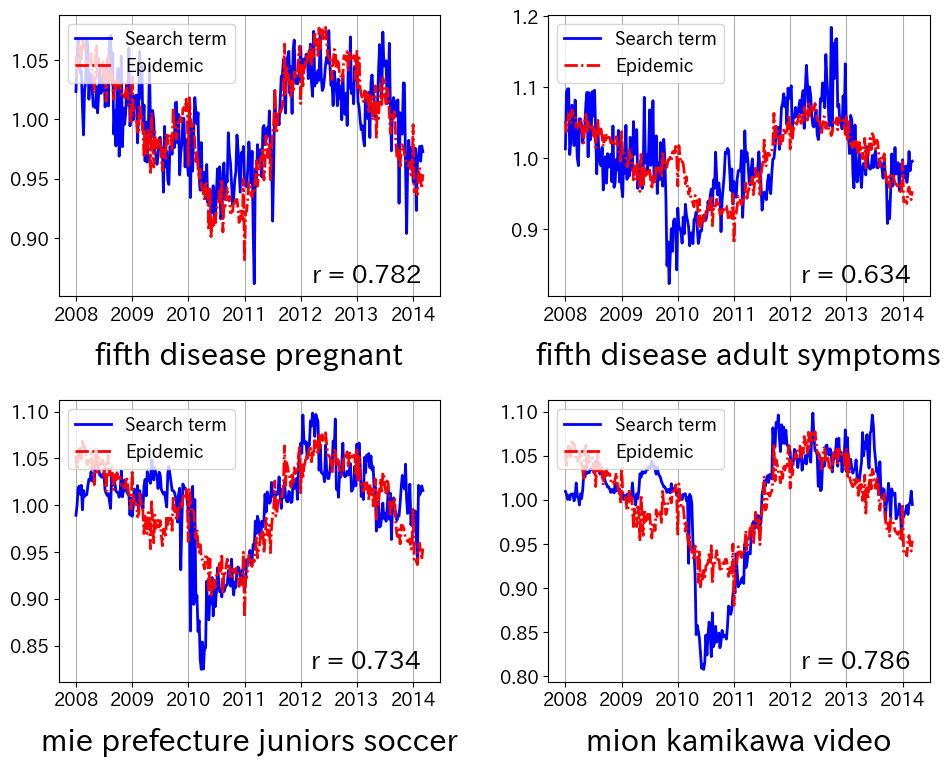}
    \caption{Irregular component: $\hat{I}$ and $\hat{I}^k$}
  \end{subfigure}%
  \vspace{-0.5em} 
  \caption{The relationship between the epidemic of erythema infectiosum and four search terms for three components.\\ We also annotate the Pearson correlation $cor(\hat{S}, \hat{S}^k)$, $cor(\Delta_1(\hat{T}),\Delta_1(\hat{T}^k))$ and $cor(\hat{I}, \hat{I}^k)$ in each subfigure. The top row and bottom row present results of search terms semantically related and non related to the target disease, respectively.}\label{fig:trend-seasonal-remainder}
 \vspace{-0.5em}
\end{figure*}

\subsubsection{\textbf{Irregular component $\hat{I}$}}\

The irregular component $\hat{I}_t$ represents the component that could not be expressed by the trend component $\hat{T}_t$ and the seasonal component $\hat{S}_t$. In other words, the irregular component represents the short-term change in the time series. The irregular component can be calculated as follows:
\begin{equation}
  \hat{I}_t = \frac{logit(y_t)}{\hat{T}_t * \hat{S}_t}.
\end{equation}
We illustrate the irregular component for the case of erythema infectiosum in Figure \ref{fig:decomposite}e.

\subsection{Feature ranking}
\ \ \ \textbf{Scoring function}: We define the scoring functions of search terms with respect to  the seasonal component $\hat{S}^k$, trend component $\hat{T}^k$ and irregular component $\hat{I}^k$ as ${Score}_{[S]}(q^k)$, ${Score}_{[T]}(q^k)$ and ${Score}_{[I]}(q^k)$, respectively. Let $cor(X,Y)$ be the Pearson correlation between $X$ and $Y$. Our scoring functions are defined as follows:
\begin{align}
  Score_{[S]}(q^k) &= max(cor(\hat{S}, \hat{S}^k), 0), \\
  Score_{[T]}(q^k) &= \max_{\varepsilon = 1}^{3} cor(\Delta_\varepsilon(\hat{T}),\Delta_\varepsilon(\hat{T}^k)) * Score_{[S]}(q^k), \\
  Score_{[I]}(q^k) &= cor(\hat{I}, \hat{I}^k) * Score_{[S]}(q^k).
\end{align}
We score $q_k$ with respect to the seasonal component by the Pearson correlation between $\hat{S}$ of infection rate and $\hat{S}^k$ of the search term $q^k$. To evaluate the trend component, we use the first-order differential time series $\Delta_\varepsilon(Y)  = \{y_{t+\varepsilon} - y_t\}_{t=1}^{T-\varepsilon}$ which is obtained by the difference of two data points. Here $\varepsilon$ is the distance between two data points. We calculate the correlation coefficients between the two differential time series $\Delta_\varepsilon(\hat{T})$ and $\Delta_\varepsilon(\hat{T}^k)$ using several different $\varepsilon$, and then adopt the highest correlation coefficient among all $\varepsilon$. The irregular component is evaluated by the correlation coefficient between the irregular component $\hat{I}$ of the infection rate and $\hat{I}^k$ of the search term $q^k$.

Moreover, we multiply the seasonal score ${Score}_{[S]}(q^k)$ to the scoring function $Score_{[T]}(q^k)$ and $Score_{[I]}(q^k)$ to exclude non-related search terms that coincidentally have high correlation with $\hat{T}$ and $\hat{I}$. In Figure \ref{fig:trend-seasonal-remainder}, we use an example about four search terms in the case of erythema infectiosum to show that multiplying the seasonal score works appropriately to select related search terms. To be specific, we consider two related search terms `fifth disease pregnant' and `fifth disease adult symptoms'. For non-related search terms, we consider `Mie prefecture juniours soccer' and `Mion Kamikawa video' (`Mion Kamikawa' is an actress name). We see that, compared to the related ones, these non-related search terms still achieve a similar or even higher correlation coefficients in the trend and irregular component, i.e., $cor(\Delta_\varepsilon(\hat{T}),\Delta_\varepsilon(\hat{T}^k))$ and $cor(\hat{I}, \hat{I}^k)$. However, because these search terms are non-related to the disease, their seasonal scores are significantly lower than the related ones. Thus, we can use the seasonal score ${Score}_{[S]}(q^k)$ to diminish the final score $Score_{[T]}(q^k)$ and $Score_{[I]}(q^k)$ of these non-related search terms. In conclusion, our scoring functions are expected to guarantee the meaningfulness of selected search terms for the prediction model.

Furthermore, because the search terms with negative correlation are not proper for epidemic prediction apparently, we set the negative scores of the seasonal component to $0$ to get rid of the case that a positive score is given by the product of two negatives.

\vspace{.5em}
\textbf{Sorting}: For the trend component, we sort the search term list in descending order by the scoring function ${Score}_{[T]}(q^k)$ and denote the sorted list as $Q_{[T]}$. Similarly, for the irregular component, we sort the search term list in descending order by the scoring function ${Score}_{[I]}(q^k)$ and denote the sorted list as  $Q_{[I]}$.

\subsection{Feature subset selection}
Next, we find the optimal feature subset from the sorted lists $Q_{[T]}$ and $Q_{[I]}$. Because we used the same feature selection algorithm and prediction model for both the trend and irregular component, here we only describe the proposed method for the case of the trend component, i.e., using $Q_{[T]}$ to predict $\hat{T}$. First, we denote the index set of selected search terms as $\Omega_{[T]}=\{\omega_1, \omega_2, \dots, \omega_p\}$ and the corresponding input matrix as $\hat{T}^{\Omega_{[T]}}=(\hat{T}^{\omega_1}$,$\hat{T}^{\omega_2}$, $\dots, \hat{T}^{\omega_p})$.

We use the multivariate regression for the prediction model:
\begin{equation}
  \hat{T}_t = f(\hat{T}^{\Omega_{[T]}}_t) = \beta_0 + \sum_{i=1}^p\beta_i \hat{T}^{\omega_i}_t.
\end{equation}

To overcome the multicollinearity problem, we learn this model with the L2 regularization. The quadratic loss function with L2 regularization is defined as follows:
\begin{equation}
  L(\hat{T}^{\Omega_{[T]}}, \hat{T}) = \frac{1}{2}\sum_t\big(\hat{T}_t - f(\hat{T}^{\Omega_{[T]}}_t)\big)^2 + \lambda\frac{1}{2}\sum_{j=1}^p \beta_j^2.
\end{equation}

We denote $\Omega_{n}$ as the index set of the selected feature subset when finished evaluating the search term $q^{k_n}$. We described our selection method as follows:
\begin{enumerate}
\item \textbf{Initiallization:} We start with the empty set $\Omega_{0} = \emptyset$.
\item \textbf{Searching direction:} We evaluate each search term in the sorted list $Q_{[T]}=\{q^{k_1}, q^{k_2},\dots,q^{k_N}\}$ in order and decide whether to add them to the prediction model.
\item \textbf{Feature subset evaluating:} We evaluate the prediction capability of a feature subset with the 5-fold cross-validation. To be specific, we learn a multivariate regression model using the training folds and use it to predict the trend component with the validation fold. The score of the feature subset will be obtained by the averaging the MSE values on five validation folds. We denote the score of the index set $\Omega$ as ${WrapperScore}_\text{PP}(\Omega)$.
\item \textbf{Feature subset updating:} Consider the case that we have to evaluate the next candidate search term $q^{k_{n+1}}$ to determine $\Omega_{n+1}$. If ${WrapperScore}_\text{PP}(\Omega_{n}\cap k_{n+1})$ is smaller than ${WrapperScore}_\text{PP}(\Omega_{n})$, we update $\Omega_{n} = \Omega_{n-1} \cap \{k_n\}$. Otherwise, we maintain $\Omega_{n} = \Omega_{n-1}$.
\item \textbf{Stop condition:} We stop adding features to $\Omega_{n}$ when the update fails five consecutive times.
\end{enumerate}

\subsection{Final prediction model}

Let $\hat{T}^{\Omega_{[T]}}$ and $\hat{I}^{\Omega_{[I]}}$ be the matrixes obtained for trend component and irregular component by feature selection. Finally, we learn two functions $f_{[T]} : \mathbb{R}^{|\Omega{[T]}|} \rightarrow \mathbb{R}$ and $f_{[I]}: \mathbb{R}^{|\Omega{[I]}|} \rightarrow \mathbb{R}$. Here $f_{[T]}$ takes $\hat{T}^{\Omega_{[T]}}_t$ as the input and returns $\hat{T}_t$. Similarly, $f_{[I]}$ takes $\hat{I}^{\Omega_{[I]}}_t$ as the input and returns $\hat{I}_t$. In other words, $f_{[T]}$ and $f_{[I]}$ predicts the trend and the irregular component, respectively. Moreover, we use the seasonal component $\hat{S}$ computed by the epidemic data of the training period as the seasonal component for the prediction model. Hence, the prediction value of infectious epidemic at $t$ is given as $y_t = logistic(\hat{S}_t * f_{[T]}(\hat{T}^{\Omega_{[T]}}_t) * f_{[I]}(\hat{I}^{\Omega_{[I]}}_t))$.

\subsection{Extension for future prediction}\label{extension}
Next, we discuss the extension to forecasting problem when $\phi\neq0$ where $\phi$ denotes the time-lag parameter. Here, we learn a function $f$ that use the data $X_{1:t}$ to predict infectious rate at time $t + \phi$, $y_{t+\phi}$.

We use the following scoring functions for the feature selection process of the forecasting problem. For the seasonal component, we use the same scoring function as when $\phi=0$. For the trend and irregular component, we first compute the time-lag time series for search rate $Y_\phi = \{y'_{t} | y'_{t+\phi} = y_{t}\}$ obtained by shifting original time series $Y$ by $\phi$ weeks. Then, we use this time-lag time series $Y_\phi$ to calculate the lagged trend $\hat{T}_\phi$, seasonal $\hat{S}_\phi$ and irregular component $\hat{I}_\phi$. We defined the scoring function as follows:
\begin{align}
  {Score}_{[S]}(q^k) &= max(cor(\hat{S}, \hat{S}^k), 0), \\
  {Score}_{[T]}(q^k) &= \max_{\varepsilon = 1}^{3} cor(\Delta_\varepsilon(\hat{T}_\phi),\Delta_\varepsilon(\hat{T}^k)) * {Score}_{[S]}(q^k), \\
  {Score}_{[I]}(q^k) &= cor(\hat{I}_\phi, \hat{I}^k) * {Score}_{[S]}(q^k).
\end{align}

The final prediction model is defined as follows:
\begin{equation}
  y_{t+\phi} = logistic\big(\hat{S}_{\phi_t} * f_{[T]}(\hat{T}^{\Omega_{[T]}}_t) * f_{[I]}(\hat{I}^{\Omega_{[I]}}_t)\big).
\end{equation}

\section{Experiment} \label{experiment}
In this section, we discuss the experiments and the results.

\subsection{Experiment data}\label{data-preparation}
\ \ \ \textbf{IDWR data:} We consider infectious diseases that the number of patients is relatively large because the clues for predicting the epidemic will hardly appear in the search logs if the number of patients is small. In our experiments, we focus on the ten infectious diseases shown in Table \ref{table:result}.

\vspace{.5em}
\textbf{Search log:} For the search log data, we use the most frequently searched 2.5 million search terms for the experiment. nIndeed, the search term ranked $2.5\times10^6$th from the top is searched roughly 450 times a day, on average.

\vspace{.5em}
\textbf{Data period:} In our experiments, we use 9 years of data from 2007 to 2015. In addition, 80\% of the dataset, which contains 374 weeks from January 01, 2007 to March 11, 2014, was used for the feature selection and training phase. The other 20\% of the data set, which contains 94 weeks from March 12, 2014, to December 31, 2015, was used for evaluating the final model.

\vspace{.5em}
\textbf{The logit function:} The search rates and infection rates used in this work are normalized within the range of $[0,1]$. We use the logit function $logit(x)=ln(\frac{x}{1-x})$ to map the rates in $[0,1]$ to $[-\infty,\infty]$. Also, because the logit function cannot be calculated when $x=0$, before applying the logit function, we replace all the data points with value $0$ with the smallest non-zero value of that time series.

\subsection{Experimental setting}\label{experiment-setting}
be evaluate the performance of the proposed method and three comparative methods on the prediction of the outbreak of ten different types of infectious diseases, with different time-lag parameters $\phi\in\{0,1,2\}$.

To evaluate the prediction accuracy, we use the Symmetric Mean Absolute Percentage Error (sMAPE) \cite{goodwin1999asymmetry} and the Pearson correlation.

Additionally, in the field of public health, the prediction is required to be made based on meaningful evidence. To evaluate the relevance of the search terms selection, we manually evaluate the interpretability of selected search terms. Specifically, we determined a search term as a related one if it contains either the target disease's name, symptoms or treatment. Moreover, we count the number of the semantically related health-care search terms listed by feature selection methods and then calculate the ratio of these related search terms to the total number of selected search terms. We call this ratio the relevance ratio of the selected search terms.

\subsection{Comparative baseline}\label{baseline}
We compare our method with three baselines, the GFT method \cite{ginsberg2009detecting}, the ElasticNet method \cite{lampos2015advances}, and the method of Neils et al. \cite{dalum2017seasonal}. First, the GFT method is evaluated with our search log data of 2.5 million search terms. For the ElasticNet method, we conducted the learning process on top 1000 search terms ranked by the GFT method. The regularization parameters were chosen using 5-fold cross-validation. Finally, for the method of Neils et al., we first applied their ranking method to rank our 2.5 million search terms and then select the best feature subset from the top 100 search terms. For the forecasting setting, we shift the infection rate time series by $\phi$ week to obtain the time-lag time series and then apply the same algorithm as the now-casting setting.

\subsection{Experiment results}\label{experiment-result}
\begin{figure*}
  \begin{subfigure}[b]{1\textwidth}
    \centering
    \begin{subfigure}[b]{0.59\textwidth}
      \centering
      \includegraphics[width=0.49\textwidth]{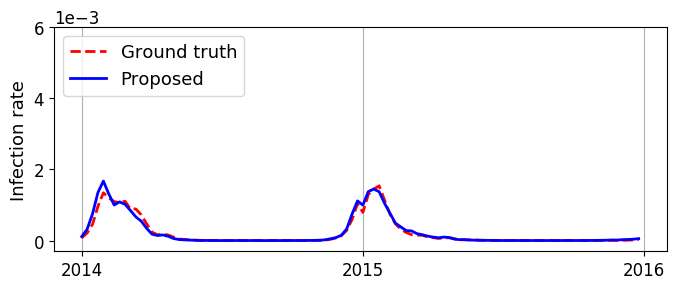}
      \includegraphics[width=0.49\textwidth]{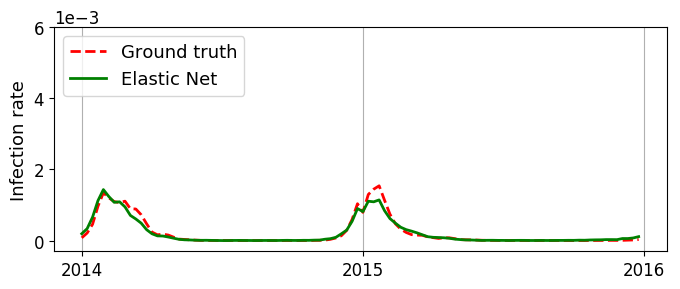}
      \includegraphics[width=0.49\textwidth]{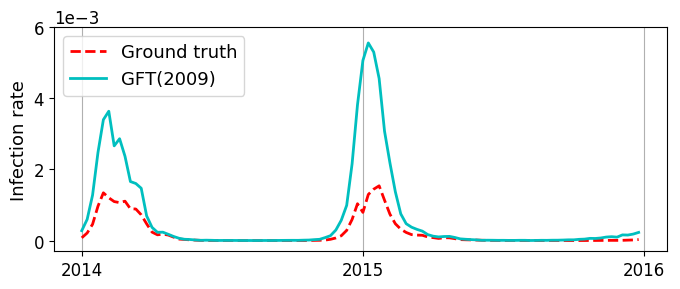}
      \includegraphics[width=0.49\textwidth]{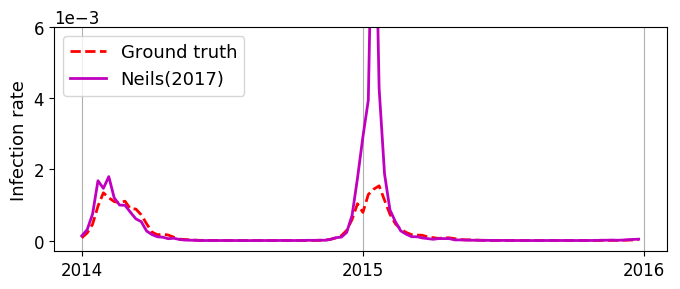}
    \end{subfigure}
    \begin{subfigure}[b]{0.39\textwidth}
      \includegraphics[width=0.95\textwidth]{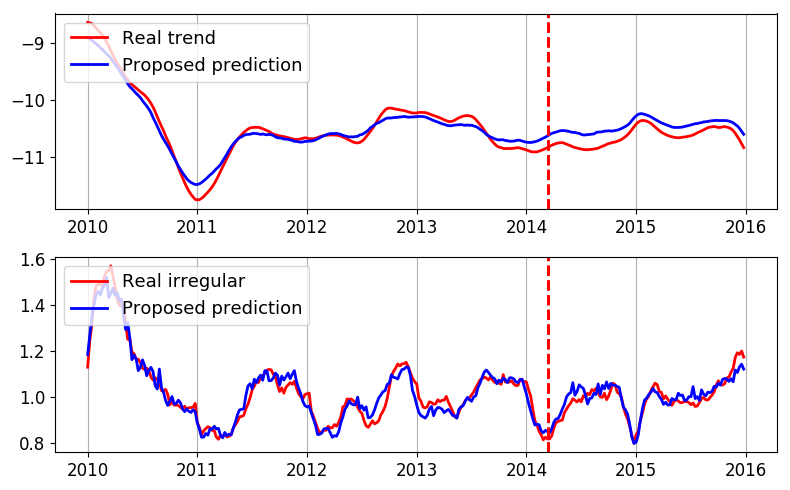}
    \end{subfigure}
    \caption{Influenza(The flu)}
    \label{fig:influ}
  \end{subfigure}
  \begin{subfigure}[b]{1\textwidth}
    \centering
    \begin{subfigure}[b]{0.59\textwidth}
      \centering
      \includegraphics[width=0.49\textwidth]{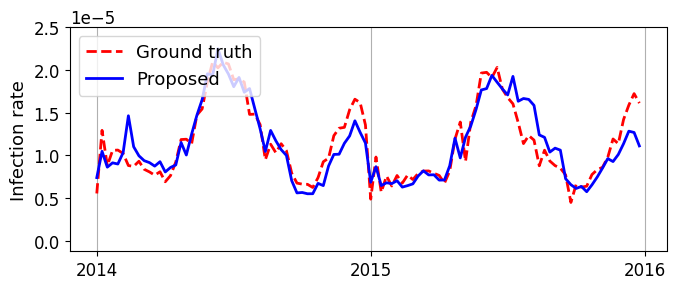}
      \includegraphics[width=0.49\textwidth]{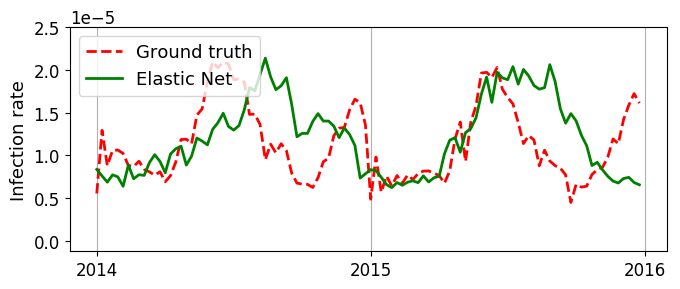}
      \includegraphics[width=0.49\textwidth]{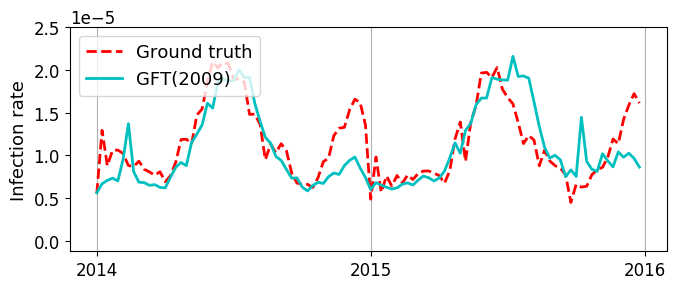}
      \includegraphics[width=0.49\textwidth]{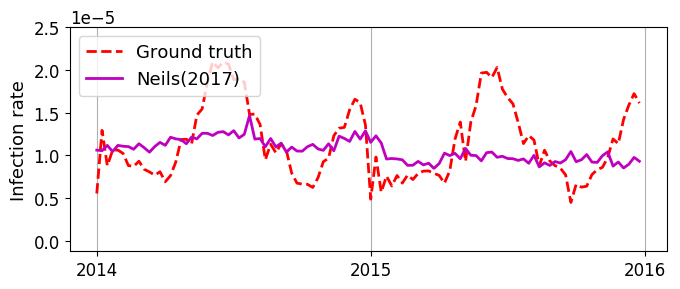}
    \end{subfigure}
    \begin{subfigure}[b]{0.39\textwidth}
      \includegraphics[width=0.95\textwidth]{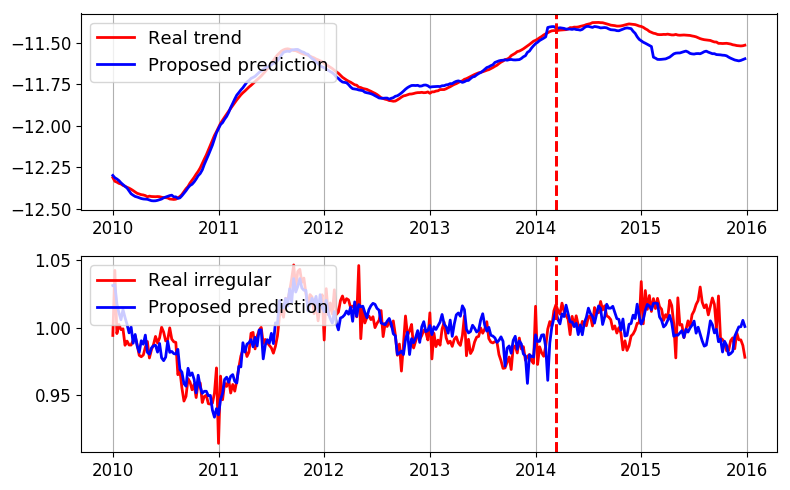}
    \end{subfigure}
    \caption{Pharyngoconjunctival fever (PCF)}
    \label{fig:pcf}
  \end{subfigure}
  \begin{subfigure}[b]{1\textwidth}
    \centering
    \begin{subfigure}[b]{0.59\textwidth}
      \centering
      \includegraphics[width=0.49\textwidth]{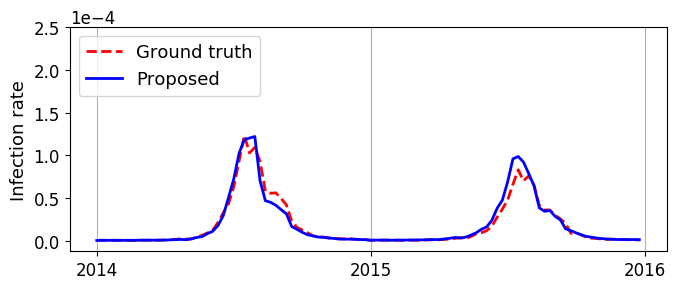}
      \includegraphics[width=0.49\textwidth]{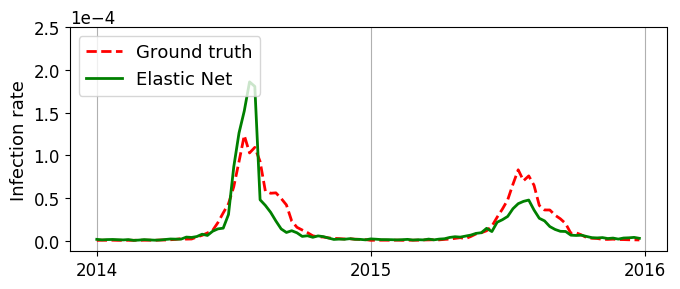}
      \includegraphics[width=0.49\textwidth]{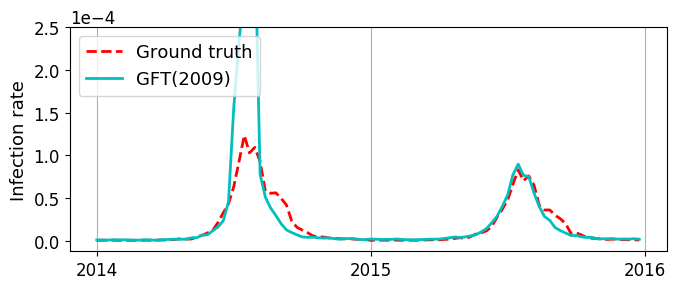}
      \includegraphics[width=0.49\textwidth]{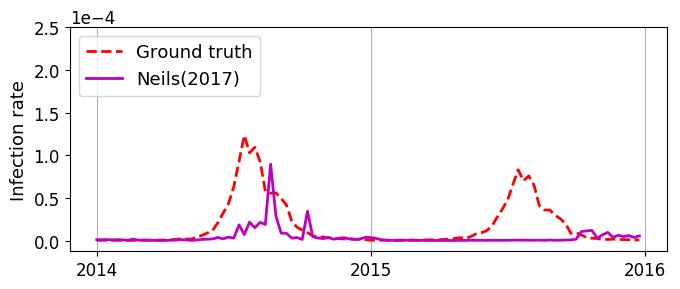}
    \end{subfigure}
    \begin{subfigure}[b]{0.39\textwidth}
      \includegraphics[width=0.95\textwidth]{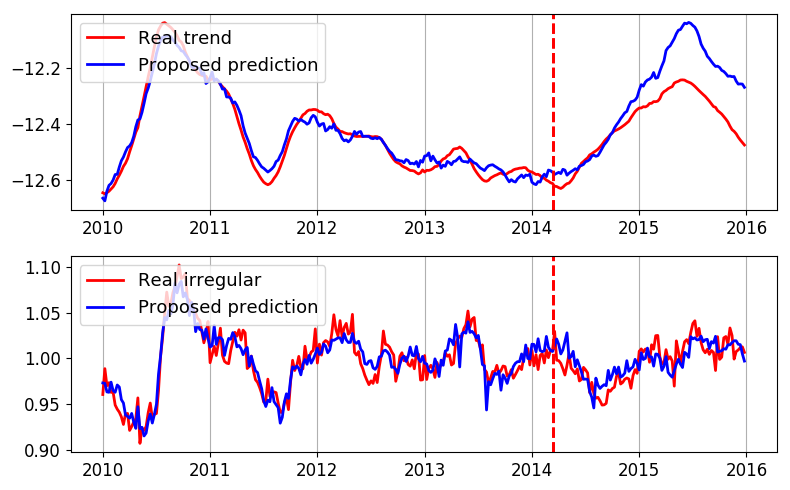}
    \end{subfigure}
    \caption{Herpangina}   
    \label{fig:herpangina}
  \end{subfigure}  
  \vspace{-1.5em}
  \caption{(Column 1 and 2) The prediction by the proposed method and three comparative methods for four infectious diseases. (Column 3) The prediction for the trend and irregular component by our method. The red vertical line indicates the test period.}
  \vspace{-0.5em}
\end{figure*}

\begin{table*}[t]
  \centering
  \resizebox{0.95\textwidth}{!}{%
    \begin{tabular}{c||rrrr||rrrr||lllll}
      \toprule
      & \multicolumn{4}{c||}{Correlation} & \multicolumn{4}{c||}{sMAPE} & \multicolumn{5}{c}{Relevant ratio} \\
\midrule
Time lag & GFT(2009) & ElasticNet & Neils(2017) & Proposed & GFT(2009) & ElasticNet & Neils(2017) & Proposed & GFT(2009) & ElasticNet & Neils(2017) & Proposed[T] & Proposed[I] \\
\midrule
\multicolumn{14}{c}{Influenza (The flu)} \\
\midrule
0 & 0.95& \textbf{0.99}& 0.77& \textbf{0.99}& 92.39& 58.47& 44.95& \textbf{30.24}& 100\% (7/7)& 33\% (9/27)& 6\% (6/99)& 100\% (28/28)& 100\% (12/12)\\
1 & \textbf{0.99}& 0.98& 0.69& 0.98& 95.37& 44.14& 42.75& \textbf{31.02}& 100\% (91/91)& 51\% (51/99)& 6\% (6/99)& 100\% (7/7)& 100\% (10/10)\\
2 & 0.93& \textbf{0.95}& 0.66& 0.93& 65.29& 42.53& 70.56& \textbf{41.80}& 100\% (4/4)& 38\% (229/589)& 4\% (4/96)& 100\% (13/13)& 100\% (4/4)\\
\midrule
\multicolumn{14}{c}{Hand, foot and mouth disease (HFMD)} \\
\midrule
0 & \textbf{0.98}& 0.97& 0.85& \textbf{0.98}& \textbf{30.32}& 32.88& 65.80& 41.69& 100\% (4/4)& 23\% (3/13)& 0\% (0/60)& 100\% (6/6)& 100\% (16/16)\\
1 & 0.95& 0.98& 0.83& \textbf{0.99}& 57.94& \textbf{32.28}& 61.39& 46.77& 50\% (2/4)& 15\% (2/13)& 0\% (0/64)& 100\% (3/3)& 88\% (23/26)\\
2 & 0.56& 0.73& 0.63& \textbf{0.97}& 78.89& 65.13& 71.74& \textbf{40.83}& 0\% (0/3)& 10\% (3/30)& 0\% (0/79)& 100\% (3/3)& 100\% (11/11)\\
\midrule
\multicolumn{14}{c}{Chickenpox (Varicella)} \\
\midrule
0 & \textbf{0.78}& 0.58& 0.37& 0.77& 43.07& 64.92& 40.95& \textbf{34.64}& 85\% (6/7)& 10\% (4/38)& 0\% (0/75)& 30\% (3/10)& 50\% (4/8)\\
1 & 0.40& \textbf{0.83}& 0.20& 0.56& 56.07& \textbf{32.02}& 44.99& 53.04& 14\% (7/47)& 16\% (2/12)& 0\% (0/4)& 18\% (2/11)& 13\% (3/22)\\
2 & 0.43& 0.29& 0.24& \textbf{0.58}& 60.37& 79.49& \textbf{44.49}& 48.87& 7\% (5/71)& 0\% (0/93)& 0\% (0/67)& 33\% (4/12)& 20\% (3/15)\\
\midrule
\multicolumn{14}{c}{Erythema infectiosum (Fifth disease)} \\
\midrule
0 & 0.76& 0.75& 0.63& \textbf{0.93}& 38.30& \textbf{30.22}& 70.79& 46.57& 62\% (10/16)& 42\% (3/7)& 5\% (3/51)& 100\% (2/2)& 88\% (8/9)\\
1 & 0.84& 0.83& 0.87& \textbf{0.93}& 29.30& \textbf{22.97}& 59.23& 78.14& 47\% (10/21)& 60\% (3/5)& 9\% (1/11)& 71\% (5/7)& 80\% (8/10)\\
2 & -0.25& 0.72& 0.82& \textbf{0.89}& 71.40& \textbf{31.93}& 66.49& 82.79& 37\% (10/27)& 60\% (3/5)& 9\% (1/11)& 66\% (4/6)& 88\% (8/9)\\
\midrule
\multicolumn{14}{c}{Pharyngoconjunctival fever (PCF)} \\
\midrule
0 & 0.76& 0.31& 0.35& \textbf{0.91}& 20.06& 34.90& 29.72& \textbf{12.96}& 58\% (10/17)& 4\% (3/73)& 0\% (0/27)& 80\% (4/5)& 44\% (4/9)\\
1 & 0.71& 0.33& 0.21& \textbf{0.92}& 26.60& 29.82& 34.26& \textbf{12.97}& 75\% (3/4)& 15\% (3/20)& 0\% (0/18)& 42\% (3/7)& 41\% (5/12)\\
2 & 0.67& 0.43& 0.26& \textbf{0.92}& 27.27& 27.09& 30.86& \textbf{13.25}& 75\% (3/4)& 8\% (2/24)& 0\% (0/35)& 33\% (2/6)& 26\% (8/30)\\
\midrule
\multicolumn{14}{c}{Herpangina (Mouth blisters)} \\
\midrule
0 & 0.83& 0.86& 0.34& \textbf{0.98}& 45.50& 50.34& 120.96& \textbf{21.07}& 85\% (6/7)& 4\% (3/70)& 0\% (0/37)& 100\% (8/8)& 55\% (11/20)\\
1 & \textbf{0.92}& \textbf{0.92}& -0.07& 0.86& 51.00& 48.89& 119.56& \textbf{39.40}& 16\% (2/12)& 3\% (1/31)& 0\% (0/11)& 100\% (8/8)& 34\% (24/70)\\
2 & 0.91& 0.92& -0.11& \textbf{0.97}& 60.90& 57.13& 112.45& \textbf{41.98}& 0\% (0/29)& 0\% (0/15)& 0\% (0/1)& 100\% (6/6)& 46\% (12/26)\\
\midrule
\multicolumn{14}{c}{Streptococcal pharyngitis (Strep throat)} \\
\midrule
0 & 0.83& \textbf{0.91}& 0.28& 0.86& 21.56& \textbf{12.66}& 40.22& 24.87& 100\% (2/2)& 40\% (2/5)& 0\% (0/33)& 41\% (7/17)& 13\% (4/29)\\
1 & 0.57& 0.24& 0.40& \textbf{0.80}& 60.95& \textbf{37.35}& 40.43& 38.83& 3\% (2/63)& 0\% (0/27)& 0\% (0/66)& 22\% (5/22)& 6\% (7/108)\\
2 & 0.56& 0.30& 0.30& \textbf{0.65}& 53.96& 33.74& \textbf{32.74}& 42.21& 2\% (2/67)& 0\% (0/84)& 0\% (0/66)& 20\% (5/24)& 9\% (4/41)\\
\midrule
\multicolumn{14}{c}{Epidemic keratoconjunctivitis (EKC)} \\
\midrule
0 & 0.55& 0.53& 0.07& \textbf{0.82}& 22.82& 28.71& 24.98& \textbf{14.59}& 100\% (2/2)& 33\% (2/6)& 0\% (0/1)& 60\% (3/5)& 40\% (8/20)\\
1 & 0.33& 0.41& -0.49& \textbf{0.56}& 22.44& 43.40& 28.04& \textbf{19.17}& 2\% (1/41)& 11\% (2/18)& 0\% (0/1)& 57\% (4/7)& 18\% (5/27)\\
2 & 0.32& 0.44& -0.45& \textbf{0.73}& 21.55& 42.57& 27.60& \textbf{15.86}& 11\% (4/36)& 3\% (2/56)& 0\% (0/1)& 100\% (3/3)& 13\% (2/15)\\
\midrule
\multicolumn{14}{c}{Gastroenteritis (Infection diarrhea)} \\
\midrule
0 & 0.87& 0.87& 0.33& \textbf{0.96}& 34.32& 17.13& 34.24& \textbf{12.73}& 100\% (40/40)& 33\% (8/24)& 0\% (0/18)& 100\% (13/13)& 100\% (20/20)\\
1 & 0.87& 0.84& 0.25& \textbf{0.95}& 39.41& 18.33& 42.37& \textbf{10.48}& 100\% (2/2)& 53\% (7/13)& 0\% (0/16)& 100\% (26/26)& 100\% (21/21)\\
2 & 0.86& 0.81& 0.02& \textbf{0.93}& 14.85& 19.35& 42.13& \textbf{12.88}& 50\% (5/10)& 29\% (5/17)& 0\% (0/70)& 100\% (20/20)& 90\% (10/11)\\
\midrule
\multicolumn{14}{c}{Mycoplasma pneumonia (Walking pneumonia)} \\
\midrule
0 & 0.88& \textbf{0.89}& -0.18& 0.83& 41.06& 35.90& 35.31& \textbf{32.13}& 26\% (13/50)& 40\% (4/10)& 0\% (0/3)& 50\% (2/4)& 12\% (1/8)\\
1 & 0.49& \textbf{0.91}& -0.14& 0.70& 34.65& 35.59& 35.66& \textbf{27.75}& 16\% (13/79)& 50\% (4/8)& 0\% (0/3)& 36\% (4/11)& 5\% (1/18)\\
2 & 0.37& 0.75& 0.00& \textbf{0.77}& 37.13& 48.76& 38.31& \textbf{35.00}& 10\% (1/10)& 27\% (3/11)& 0\% (0/11)& 33\% (1/3)& 3\% (1/28)\\
\midrule
      \bottomrule
    \end{tabular}}
  \vspace{0.5em}
  \caption{The correlation coefficient (Column 1) , the sMAPE value (Column 2) and the relevance ratio with the actual number of related search terms among the selected ones (Column 3) of the proposed method and three comparative methods.}
  \label{table:result}
  \vspace{-2em}
\end{table*}
\subsubsection{\textbf{Was the prediction accuracy improved ?}}\

In the Table \ref{table:result}, we show the Pearson correlation values and the sMAPE values in the test period for each disease with different time-lag parameters $\phi$.

We first explain the results of now-casting, i.e, when $\phi = 0$. We see that our proposed method achieved higher correlation coefficients compared to all the other methods for seven of ten diseases. For the remaining three infectious diseases: chickenpox, strep throat, mycoplasma pneumonia, the proposed method could not achieve the best correlation coefficient while it still attained reasonable prediction with correlation coefficients that are close to the best one of the comparative methods. In contrast, the GFT method and the ElasticNet method only achieved the best correlation coefficient for only two of ten diseases. A similar observation can also be obtained for the sMAPE value.

From Table \ref{table:result}, we also see that the proposed method also achieved lower sMAPE values and higher correlation coefficients for the forecast problem, i.e., when $\phi\ne0$. When the time lag parameter becomes larger, the accuracy of the comparison methods decreases drastically. Though this decrease of accuracy is still observed for our method, the decreasing is much slower compared to the other methods. In particular, for the case of HFMD, Erythema infectiosum, PCF and Gastroenteritis, our proposed method achieved the highest correlation coefficient for all the time-lag parameters $\phi$.

We show the prediction results when $\phi=0$ for four different diseases: influenza (Figure \ref{fig:influ}), PCF disease (Figure \ref{fig:pcf}), herpangina (Figure \ref{fig:herpangina}). We see that, for all these diseases, the proposed method performed more accurately and stablely compared to the other methods. To be specific, for the GFT method, we observed the long-term misprediction in the case of influenza and herpangina. We also observed the short-term misprediction in the case of PCF disease. Compared to the GFT method, the ElasticNet method attained more accurate predictions. However, we observed that ElasticNet still gives the long-term mispredictions for herpangina and PCF disease. In contrast, we did not observe such kinds of mispredictions in the predictions by the proposed method. In other words, the proposed method is significantly more stable than the existing methods. We also showed the trend component and irregular component prediction of these diseases in Figure \ref{fig:influ}, \ref{fig:pcf}, and \ref{fig:herpangina}. We see that our method succeeded in predicting both the trend and irregular component for these diseases.

\subsubsection{\textbf{Which component is harder to predict?}} \label{difficult-component}
\
\begin{table}[h!]
  \centering
  \resizebox{0.45\textwidth}{!}{%
    \begin{tabular}{c|p{1.7cm}p{1.7cm}p{1.7cm}p{1.7cm}}
      \toprule
      Rank & Herpangina & Influenza & Fifth disease & HFMD \\
      \midrule
      1 & 0.83& 0.95 & 0.91 & 0.93 \\
      2 & 0.68& 0.94 & 0.85 & 0.90 \\
      3 & 0.63& 0.94 & 0.91 & 0.86\\
      4 & 0.62& 0.93 & 0.95 & 0.85\\
      5 & 0.63& 0.93 & 0.93 & 0.88\\
      \bottomrule
    \end{tabular}}
  \caption{The trend correlation $\max_{\varepsilon = 1}^{3} cor(\Delta_\varepsilon(\hat{T_\phi}),\Delta_\varepsilon(\hat{T}^k))$ of 5 top-rank search terms for the trend component}\label{table:trend}
 \vspace{-3em} 
\end{table}

From Figure \ref{fig:herpangina}, we see that the proposed method showed some overestimation and underestimation in predicting the trend of herpangina in the test period. We show Table \ref{table:trend} to discuss this failure. In this table, we show the correlations of the top-5 search terms of the trend component in the case of four diseases, including herpangina. We observed that the correlation coefficients of the trend component $\max_{\varepsilon = 1}^{3}cor(\Delta_\varepsilon(\hat{T_\phi}),\Delta_\varepsilon(\hat{T}^k))$ in the case of herpangina are significantly weaker than those of the other three diseases. Hence, in the case of herpangina, even for the top-rank search terms, the trend change of the search terms are incapable of reflecting the trend change of the epidemic. Thus, it is difficult to predict the trend of herpangina using the search terms.

Compared to herpangina, we observed that the proposed method can predict more accurately for the trend component of the other infectious diseases. However, predicting the trend component is still much harder than predicting the irregular component. As confirmed in Figure \ref{fig:herpangina}, the prediction error of the trend component tends to increases with time. To be precise, for many infectious diseases, the prediction error of the trend component in the second year of the test period is often greater than that of the first year. In contrast, we did not observe such difference in the prediction accuracy of the irregular component. 

\subsubsection{\textbf{Can we select meaningful search terms?}}\

In Table \ref{table:result}, we also show the relevance ratio and the actual number of search terms semantically related to the target diseases among the selected search terms. For example, for the now-casting setting of erythema infectiosum, 17 search terms were selected by the GFT method, in which 10 of them were related to erythema infectiosum. We express these values as `62\% (10/16)'. We call the value `58\%' the relevance ratio of the selected features. In addition, we also show some of the search terms selected by our method in the appendix.

We see that the proposed method was able to achieve better relevance ratio than the comparative methods. In particular, for the case of the now-casting problem, the proposed and the GFT method achieved the best relevance ratio among all methods. However, when $\phi$ is large, the GFT method almost failed to select semantically related search terms. In contrast, the proposed method can select semantically related search terms even when $\phi$ is large. The ElasticNet method was able to learn a sparse model for most of the cases, whereas many selected search terms are non-related to the target diseases. Furthermore, the method of Neils et al. completely failed in selecting semantically related search terms for all the diseases. The reason is that their selecting rule does not work well when applied to large-scale search logs. To be specific, because of the enormous diversity of the search terms, Neils et al.'s method selected many search terms that are highly correlated to the residual of the seasonal model, but non-related to the target disease. This failure in selection explains the poor prediction accuracy of this method shown in Table \ref{table:result}.

For some diseases, such as chickenpox, streptococcal pharyngitis, EKC and mycoplasma pneumonia, our method selected many non-related search terms for the irregular component. The first reason is that the irregular component is often noisy. In addition, the irregular component of these infectious diseases fluctuates tightly around $1.0$. As a consequence, it is difficult to distinguish the related search terms from the non-related ones. Fortunately, because the irregular components fluctuate tightly around $1.0$, they can merely be treated as noise. Thus, predicting the irregular component becomes relatively less important in the overall epidemic prediction. Even so, the proposed method still performed acceptably in predicting the irregular component of these diseases.

\subsubsection{\textbf{What kinds of search terms were selected?}}\
\begin{figure}[t]
 \centering
 \includegraphics[width=0.4\textwidth]{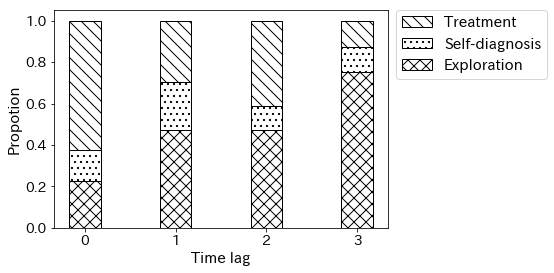}
 \vspace{-1em}   
 \caption{Categories of selected search term for influenza} 
 \label{fig:influ-selected}
 \vspace{-2em}  
\end{figure}

Next, we focus on the case of influenza to analyze the meaning the search terms selected by our method. Borrowing the ideas from existing studies on web searching behavior \cite{cartright2011intentions}, we categorize selected search terms into three categories: self-diagnosis search term, treatment search term, and exploration search term. To be specific, self-diagnosis search term is the search term that contains symptom keyword, such as `influenza fever' or `influenza headache'; treatment search term is the search term about treatments or concerns of infected users, such as `influenza medicine` or `influenza going to school'. These two categories are considered to be searched when users actually infected. Finally, exploration search term is the kind of search term that can be searched when the user is whether infected or not, such as `influenza infectiousness', `influenza epidemic'. We assume that such self-diagnosis and treatment search terms are more efficient than exploration search terms for predicting the outbreak.

In Figure \ref{fig:influ-selected}, we show the proportion of these categories for each time-lag $\phi$ in the case of influenza. Here we summed up the number of selected search terms for trend and irregular components to calculate the proportion. We see that the larger the $\phi$, the greater the proportion of exploration search terms becomes. In other words, the larger the $\phi$, the smaller the proportion of the search terms that are efficient for predicting the outbreak becomes. This is because the symptoms of influenza typically last only five to seven days for the majority of persons \cite{heidi2016flu}. This decreasing in the number of efficient search terms also well explains the decrease of the prediction accuracy in the case of influenza shown in Table \ref{table:result}.

Due to the randomness of the feature selection process, this observation might not generalize to other infectious diseases. Another insight is that both methods failed to select related search terms for pertussis and mycoplasma pneumonia. We note that these infectious diseases both have comparatively few patients and weak seasonality. Hence, we conclude that a reasonable number of patients and strong seasonality are necessary for the epidemic prediction using search logs.

\section{Conclusions}
In this work, we proposed a novel scalable feature selection method and a prediction modeling framework based on seasonal adjustment in the infectious epidemic prediction problem using search engine logs. By considering each component individually, we achieved more accurate and stable outbreak prediction against both the short-term and long-term changes in search engine logs. We showed that our proposed method is significantly outperformed the existing methods by conducting comprehensive experiments on predicting the outbreak of ten kinds of infectious diseases. The experimental results showed that the proposed method can give accurate predictions for both now-casting and forecasting setting. Moreover, the proposed method is more successful in selecting search terms semantically related to target diseases.

\begin{acks}
This work was partly supported by KAKENHI (Grants-in-Aid for scientific research) Grant Numbers JP19H04164 and JP18H04099. We would also like to thank Yahoo! JAPAN Corporation for providing the search engine logs data for this work.
\end{acks}

\bibliography{reference_ready.bib}


\begin{thebibliography}{18}

%%% the Web address.

\ifx \showCODEN    \undefined \def \showCODEN     #1{\unskip}     \fi
\ifx \showDOI      \undefined \def \showDOI       #1{#1}\fi
\ifx \showISBNx    \undefined \def \showISBNx     #1{\unskip}     \fi
\ifx \showISBNxiii \undefined \def \showISBNxiii  #1{\unskip}     \fi
\ifx \showISSN     \undefined \def \showISSN      #1{\unskip}     \fi
\ifx \showLCCN     \undefined \def \showLCCN      #1{\unskip}     \fi
\ifx \shownote     \undefined \def \shownote      #1{#1}          \fi
\ifx \showarticletitle \undefined \def \showarticletitle #1{#1}   \fi
\ifx \showURL      \undefined \def \showURL       {\relax}        \fi
% invisible to TeX
\providecommand\bibfield[2]{#2}
\providecommand\bibinfo[2]{#2}
\providecommand\natexlab[1]{#1}
\providecommand\showeprint[2][]{arXiv:#2}

\bibitem[\protect\citeauthoryear{Anderson and Perrin}{Anderson and
  Perrin}{2017}]%
        {monica2017tech}
\bibfield{author}{\bibinfo{person}{Monica Anderson} {and}
  \bibinfo{person}{Andrew Perrin}.} \bibinfo{year}{2017}\natexlab{}.
\newblock \showarticletitle{Tech Adoption Climbs Among Older Adults}.
\newblock  (\bibinfo{year}{2017}).
\newblock


\bibitem[\protect\citeauthoryear{Brownstein, Freifeld, and Madoff}{Brownstein
  et~al\mbox{.}}{2009}]%
        {brownstein2009digital}
\bibfield{author}{\bibinfo{person}{John~S Brownstein}, \bibinfo{person}{Clark~C
  Freifeld}, {and} \bibinfo{person}{Lawrence~C Madoff}.}
  \bibinfo{year}{2009}\natexlab{}.
\newblock \showarticletitle{Digital disease detection—harnessing the Web for
  public health surveillance}.
\newblock \bibinfo{journal}{\emph{New England Journal of Medicine}}
  \bibinfo{volume}{360}, \bibinfo{number}{21} (\bibinfo{year}{2009}),
  \bibinfo{pages}{2153--2157}.
\newblock


\bibitem[\protect\citeauthoryear{Butler}{Butler}{2013}]%
        {butler2013google}
\bibfield{author}{\bibinfo{person}{Declan Butler}.}
  \bibinfo{year}{2013}\natexlab{}.
\newblock \showarticletitle{When Google got flu wrong}.
\newblock \bibinfo{journal}{\emph{Nature}} \bibinfo{volume}{494},
  \bibinfo{number}{7436} (\bibinfo{year}{2013}), \bibinfo{pages}{155}.
\newblock


\bibitem[\protect\citeauthoryear{Cartright, White, and Horvitz}{Cartright
  et~al\mbox{.}}{2011}]%
        {cartright2011intentions}
\bibfield{author}{\bibinfo{person}{Marc-Allen Cartright},
  \bibinfo{person}{Ryen~W White}, {and} \bibinfo{person}{Eric Horvitz}.}
  \bibinfo{year}{2011}\natexlab{}.
\newblock \showarticletitle{Intentions and attention in exploratory health
  search}. In \bibinfo{booktitle}{\emph{Proceedings of the 34th international
  ACM SIGIR conference on Research and development in Information Retrieval}}.
  ACM, \bibinfo{pages}{65--74}.
\newblock


\bibitem[\protect\citeauthoryear{Copeland, Romano, Zhang, Hecht, Zigmond, and
  Stefansen}{Copeland et~al\mbox{.}}{2013}]%
        {copeland2013google}
\bibfield{author}{\bibinfo{person}{Patrick Copeland}, \bibinfo{person}{Raquel
  Romano}, \bibinfo{person}{Tom Zhang}, \bibinfo{person}{Greg Hecht},
  \bibinfo{person}{Dan Zigmond}, {and} \bibinfo{person}{Christian Stefansen}.}
  \bibinfo{year}{2013}\natexlab{}.
\newblock \showarticletitle{Google disease trends: an update}.
\newblock \bibinfo{journal}{\emph{Nature}}  \bibinfo{volume}{457}
  (\bibinfo{year}{2013}), \bibinfo{pages}{1012--1014}.
\newblock


\bibitem[\protect\citeauthoryear{Dalum~Hansen, M{\o}lbak, Cox, and
  Lioma}{Dalum~Hansen et~al\mbox{.}}{2017}]%
        {dalum2017seasonal}
\bibfield{author}{\bibinfo{person}{Niels Dalum~Hansen},
  \bibinfo{person}{K{\aa}re M{\o}lbak}, \bibinfo{person}{Ingemar~J Cox}, {and}
  \bibinfo{person}{Christina Lioma}.} \bibinfo{year}{2017}\natexlab{}.
\newblock \showarticletitle{Seasonal Web Search Query Selection for
  Influenza-Like Illness (ILI) Estimation}. In
  \bibinfo{booktitle}{\emph{Proceedings of the 40th International ACM SIGIR
  Conference on Research and Development in Information Retrieval}}. ACM,
  \bibinfo{pages}{1197--1200}.
\newblock


\bibitem[\protect\citeauthoryear{Ginsberg, Mohebbi, Patel, Brammer, Smolinski,
  and Brilliant}{Ginsberg et~al\mbox{.}}{2009}]%
        {ginsberg2009detecting}
\bibfield{author}{\bibinfo{person}{Jeremy Ginsberg}, \bibinfo{person}{Matthew~H
  Mohebbi}, \bibinfo{person}{Rajan~S Patel}, \bibinfo{person}{Lynnette
  Brammer}, \bibinfo{person}{Mark~S Smolinski}, {and} \bibinfo{person}{Larry
  Brilliant}.} \bibinfo{year}{2009}\natexlab{}.
\newblock \showarticletitle{Detecting influenza epidemics using search engine
  query data}.
\newblock \bibinfo{journal}{\emph{Nature}} \bibinfo{volume}{457},
  \bibinfo{number}{7232} (\bibinfo{year}{2009}), \bibinfo{pages}{1012}.
\newblock


\bibitem[\protect\citeauthoryear{Godman}{Godman}{2016}]%
        {heidi2016flu}
\bibfield{author}{\bibinfo{person}{Heidi Godman}.}
  \bibinfo{year}{2016}\natexlab{}.
\newblock \showarticletitle{How long does the flu last?}
\newblock  (\bibinfo{year}{2016}).
\newblock


\bibitem[\protect\citeauthoryear{Goodwin and Lawton}{Goodwin and
  Lawton}{1999}]%
        {goodwin1999asymmetry}
\bibfield{author}{\bibinfo{person}{Paul Goodwin} {and} \bibinfo{person}{Richard
  Lawton}.} \bibinfo{year}{1999}\natexlab{}.
\newblock \showarticletitle{On the asymmetry of the symmetric MAPE}.
\newblock \bibinfo{journal}{\emph{International journal of forecasting}}
  \bibinfo{volume}{15}, \bibinfo{number}{4} (\bibinfo{year}{1999}),
  \bibinfo{pages}{405--408}.
\newblock


\bibitem[\protect\citeauthoryear{Guyon and Elisseeff}{Guyon and
  Elisseeff}{2003}]%
        {guyon2003introduction}
\bibfield{author}{\bibinfo{person}{Isabelle Guyon} {and}
  \bibinfo{person}{Andr{\'e} Elisseeff}.} \bibinfo{year}{2003}\natexlab{}.
\newblock \showarticletitle{An introduction to variable and feature selection}.
\newblock \bibinfo{journal}{\emph{Journal of machine learning research}}
  \bibinfo{volume}{3}, \bibinfo{number}{Mar} (\bibinfo{year}{2003}),
  \bibinfo{pages}{1157--1182}.
\newblock


\bibitem[\protect\citeauthoryear{(JAPAN)}{(JAPAN)}{[n.d.]}]%
        {idwr-data}
\bibfield{author}{\bibinfo{person}{National Institute Of Infectious~Diseases
  (JAPAN)}.} \bibinfo{year}{[n.d.]}\natexlab{}.
\newblock \bibinfo{title}{Infectious Diseases Weekly Report (IDWR)}.
\newblock \bibinfo{howpublished}{https://www.niid.go.jp/niid/en/idwr-e.html}.
\newblock


\bibitem[\protect\citeauthoryear{Kohavi and John}{Kohavi and John}{1997}]%
        {kohavi1997wrappers}
\bibfield{author}{\bibinfo{person}{Ron Kohavi} {and} \bibinfo{person}{George~H
  John}.} \bibinfo{year}{1997}\natexlab{}.
\newblock \showarticletitle{Wrappers for feature subset selection}.
\newblock \bibinfo{journal}{\emph{Artificial intelligence}}
  \bibinfo{volume}{97}, \bibinfo{number}{1-2} (\bibinfo{year}{1997}),
  \bibinfo{pages}{273--324}.
\newblock


\bibitem[\protect\citeauthoryear{Lampos, Miller, Crossan, and Stefansen}{Lampos
  et~al\mbox{.}}{2015}]%
        {lampos2015advances}
\bibfield{author}{\bibinfo{person}{Vasileios Lampos}, \bibinfo{person}{Andrew~C
  Miller}, \bibinfo{person}{Steve Crossan}, {and} \bibinfo{person}{Christian
  Stefansen}.} \bibinfo{year}{2015}\natexlab{}.
\newblock \showarticletitle{Advances in nowcasting influenza-like illness rates
  using search query logs}.
\newblock \bibinfo{journal}{\emph{Scientific reports}}  \bibinfo{volume}{5}
  (\bibinfo{year}{2015}), \bibinfo{pages}{12760}.
\newblock


\bibitem[\protect\citeauthoryear{Lampos, Zou, and Cox}{Lampos
  et~al\mbox{.}}{2017}]%
        {lampos2017enhancing}
\bibfield{author}{\bibinfo{person}{Vasileios Lampos}, \bibinfo{person}{Bin
  Zou}, {and} \bibinfo{person}{Ingemar~Johansson Cox}.}
  \bibinfo{year}{2017}\natexlab{}.
\newblock \showarticletitle{Enhancing feature selection using word embeddings:
  The case of flu surveillance}. In \bibinfo{booktitle}{\emph{Proceedings of
  the 26th International Conference on World Wide Web}}. International World
  Wide Web Conferences Steering Committee, \bibinfo{pages}{695--704}.
\newblock


\bibitem[\protect\citeauthoryear{Polgreen, Chen, Pennock, Nelson, and
  Weinstein}{Polgreen et~al\mbox{.}}{2008}]%
        {polgreen2008using}
\bibfield{author}{\bibinfo{person}{Philip~M Polgreen}, \bibinfo{person}{Yiling
  Chen}, \bibinfo{person}{David~M Pennock}, \bibinfo{person}{Forrest~D Nelson},
  {and} \bibinfo{person}{Robert~A Weinstein}.} \bibinfo{year}{2008}\natexlab{}.
\newblock \showarticletitle{Using internet searches for influenza
  surveillance}.
\newblock \bibinfo{journal}{\emph{Clinical infectious diseases}}
  \bibinfo{volume}{47}, \bibinfo{number}{11} (\bibinfo{year}{2008}),
  \bibinfo{pages}{1443--1448}.
\newblock


\bibitem[\protect\citeauthoryear{Tibshirani}{Tibshirani}{1996}]%
        {tibshirani1996regression}
\bibfield{author}{\bibinfo{person}{Robert Tibshirani}.}
  \bibinfo{year}{1996}\natexlab{}.
\newblock \showarticletitle{Regression shrinkage and selection via the lasso}.
\newblock \bibinfo{journal}{\emph{Journal of the Royal Statistical Society.
  Series B (Methodological)}} (\bibinfo{year}{1996}),
  \bibinfo{pages}{267--288}.
\newblock


\bibitem[\protect\citeauthoryear{Yuan, Nsoesie, Lv, Peng, Chunara, and
  Brownstein}{Yuan et~al\mbox{.}}{2013}]%
        {yuan2013monitoring}
\bibfield{author}{\bibinfo{person}{Qingyu Yuan}, \bibinfo{person}{Elaine~O
  Nsoesie}, \bibinfo{person}{Benfu Lv}, \bibinfo{person}{Geng Peng},
  \bibinfo{person}{Rumi Chunara}, {and} \bibinfo{person}{John~S Brownstein}.}
  \bibinfo{year}{2013}\natexlab{}.
\newblock \showarticletitle{Monitoring influenza epidemics in china with search
  query from baidu}.
\newblock \bibinfo{journal}{\emph{PloS one}} \bibinfo{volume}{8},
  \bibinfo{number}{5} (\bibinfo{year}{2013}), \bibinfo{pages}{e64323}.
\newblock


\bibitem[\protect\citeauthoryear{Zou and Hastie}{Zou and Hastie}{2005}]%
        {zou2005regularization}
\bibfield{author}{\bibinfo{person}{Hui Zou} {and} \bibinfo{person}{Trevor
  Hastie}.} \bibinfo{year}{2005}\natexlab{}.
\newblock \showarticletitle{Regularization and variable selection via the
  elastic net}.
\newblock \bibinfo{journal}{\emph{Journal of the Royal Statistical Society:
  Series B (Statistical Methodology)}} \bibinfo{volume}{67},
  \bibinfo{number}{2} (\bibinfo{year}{2005}), \bibinfo{pages}{301--320}.
\newblock


\end{thebibliography}

\appendix
\section{Selected search terms}
Here we show some of top selected search terms by the proposed method for the now-casting task (originally in Japanese)

\begin{table}[b]
 \centering
 \begin{tabular}{l|l}
 
 \multicolumn{2}{c}{Influenza (The flu)} \\ \hline \hline
 Trend component & Irregular component\\ \hline  
  influenza incubation period & influenza incubation period\\
 influenza infection period & influenza infection period\\
 after influenza fever & tamiflu side effect\\
 tamiflu side effect & influenza recovery\\
 influenza recovery & influenza school\\
 school of influenza & influenza fever\\
 influenza antipyretic & influenza fever\\
 influenza going out & influenza antipyretics\\
 influenza work & influenza a type\\
  \hline

  \multicolumn{2}{c}{} \\
  \multicolumn{2}{c}{Hand, foot and mouth disease (HFMD)} \\ \hline \hline
  Trend component & Irregular component\\ \hline  
  HFMD pregnant women& HFMD disease \\
  HFMD (Hand Foot Mouth)& HFMD pregnant women \\
  HMFD (Hand Mouth Foot)& HFMD adult \\
  adults HFMD & HFMD treatment \\
  HFMD stomatitis & HFMD infection \\
  HFMD newborn & adults HFMD \\
  & HFMD medicine \\
  & HFMD fever \\
  & high fever HFMD \\
   \hline
  
 \multicolumn{2}{c}{} \\
 \multicolumn{2}{c}{Chickenpox (Varicella)} \\ \hline \hline
 Trend component & Irregular component\\ \hline    
 chickenpox (in Kanji) & chickenpox (in Kanji)  \\
 chickenpox (in Hiragana) & chickenpox (in Hiragana) \\
 ikimonogakari thank you lyrics & chickenpox pregnant woman \\
 http://www.kuronekoy & http://www.kuronekoy \\
 first love forecast (Comic) & 0471792338 \\
 doltz & playstation 3 torne \\ 
 electric toothbrush & ikimonogakari thank you lyrics \\ 
 chickenbox (in Romaji) & Liquefied Phenol \\
 gem night deck (Card Game) & \\
 \hline 

 \multicolumn{2}{c}{} \\
 \multicolumn{2}{c}{Erythema infectiosum (Fifth disease)} \\ \hline \hline
 Trend component & Irregular component\\ \hline    
  adult fifth disease & fifth disease adults \\
  fifth disease & fifth disease (in Hiragana) \\
  & erythema infectious \\
  & fifth disaese pregnant women \\
  & fifth disease incubation period \\
  & adults fifth disease \\
  & fifth disease (in Katakana) \\
  & skype bulletin board \\
  & unisia combination tablet \\
 \hline  
 \end{tabular}
\end{table}

\begin{table}[b]
 \vspace{6.3em} 
 \centering
 \begin{tabular}{l|l}
  \multicolumn{2}{c}{Pharyngoconjunctival fever (PCF)} \\ \hline \hline
 Trend component & Irregular component\\ \hline  
 PCF fever & PCF fever\\
 pharyngeal conjunctival fever & riders jacket\\
 pharyngeal conjunctivitis & adeno virus\\
 tamiflu side effect & adeno\\
 riders jacket & hachioji castle\\
 hachioji castle & PCF disease\\
 adeno & kuwabata blog \\
 & epidemic conjunctivitis\\
 & bromidrosis\\
 \hline

  \multicolumn{2}{c}{} \\
  \multicolumn{2}{c}{Hand, foot and mouth disease (HFMD)} \\ \hline \hline
  Trend component & Irregular component\\ \hline  
  HFMD pregnant women& HFMD disease \\
  HFMD (Hand Foot Mouth)& HFMD pregnant women \\
  HMFD (Hand Mouth Foot)& HFMD adult \\
  adults HFMD & HFMD treatment \\
  HFMD stomatitis & HFMD infection \\
  HFMD newborn & adults HFMD \\
  & HFMD medicine \\
  & HFMD fever \\
  & high fever HFMD \\
   \hline
  
 \multicolumn{2}{c}{} \\
 \multicolumn{2}{c}{Herpangina (Mouth blisters)} \\ \hline \hline
 Trend component & Irregular component\\ \hline
helpangina & helpangina \\
herpangina pregnant woman & HFMD \\
helpangina incubation period & helpangina incubation period \\
HFMD disease & helpangina adult \\
helpengina (typo) & HFMD prevention \\
HFMD epidemic & HFMD food \\
helbangina (typo) & helpangina medicine \\
HFMD adult & helpangina symptoms \\
& aiko set list (singer show ticket) \\
\hline 

 \multicolumn{2}{c}{} \\
 \multicolumn{2}{c}{Gastroenteritis (Infection diarrhea)} \\ \hline \hline
 Trend component & Irregular component\\ \hline
viral gastroenteritis & viral gastroenteritis\\
infectious disease gastroenteritis & children vomiting\\
children vomiting & infectious disease gastroenteritis\\
gastrointestinal cold & infant vomitting\\
epidemic gastroenteritis & infectious gastroenteritis\\
gastroenteritis & vomiting diarrhea\\
baby gastroenteritis & gastroenteritis\\
diarrhea vomiting & epidemic emesis diarrhea\\
vomiting diarrhea & child gastroenteritis\\
 \hline  
 \end{tabular}
\end{table} 

\end{document}